\documentclass{article}
\usepackage{spconf,graphicx}

\usepackage[]{amsmath, amssymb}
\usepackage[]{csquotes}
\usepackage{xcolor}
\def\ie{{i.e.\,}}

\def\rms{\vspace{-.2cm}}
\DeclareMathOperator*{\argmax}{arg\,max}

\title{Cell segmentation with random ferns and graph-cuts}
\name{A. Browet$^{\star}$, C. De Vleeschouwer$^{\star}$ , L. Jacques$^{\star}$, N. Mathiah$^{\dagger}$, B. Saykali$^{\dagger}$, I. Migeotte$^{\dagger}$ 
\thanks{$\dagger$ Part of this work was supported by a fellowship from Phoenix, the FRS/FRIA, the FNRS, WELBIO and by the Mecatech project SAVE.}%Thanks to financial support $\Rightarrow$ Part of this work has been funded by the Belgian N.S.F., and by the
}

\address{$^{\star}$Universit\'e catholique de Louvain, ICTEAM Institute, Belgium.\\
$^{\dagger}$Universit\'e Libre de Bruxelles, IRIBHM, Belgium\\}
\def\imgwidth{1\linewidth}
\def\minipagewidth{.185\linewidth}
\def\includelargeimage{
\begin{figure*}[t]
\begin{minipage}[b]{\minipagewidth}
  \centering
  \centerline{\includegraphics[width=\imgwidth]{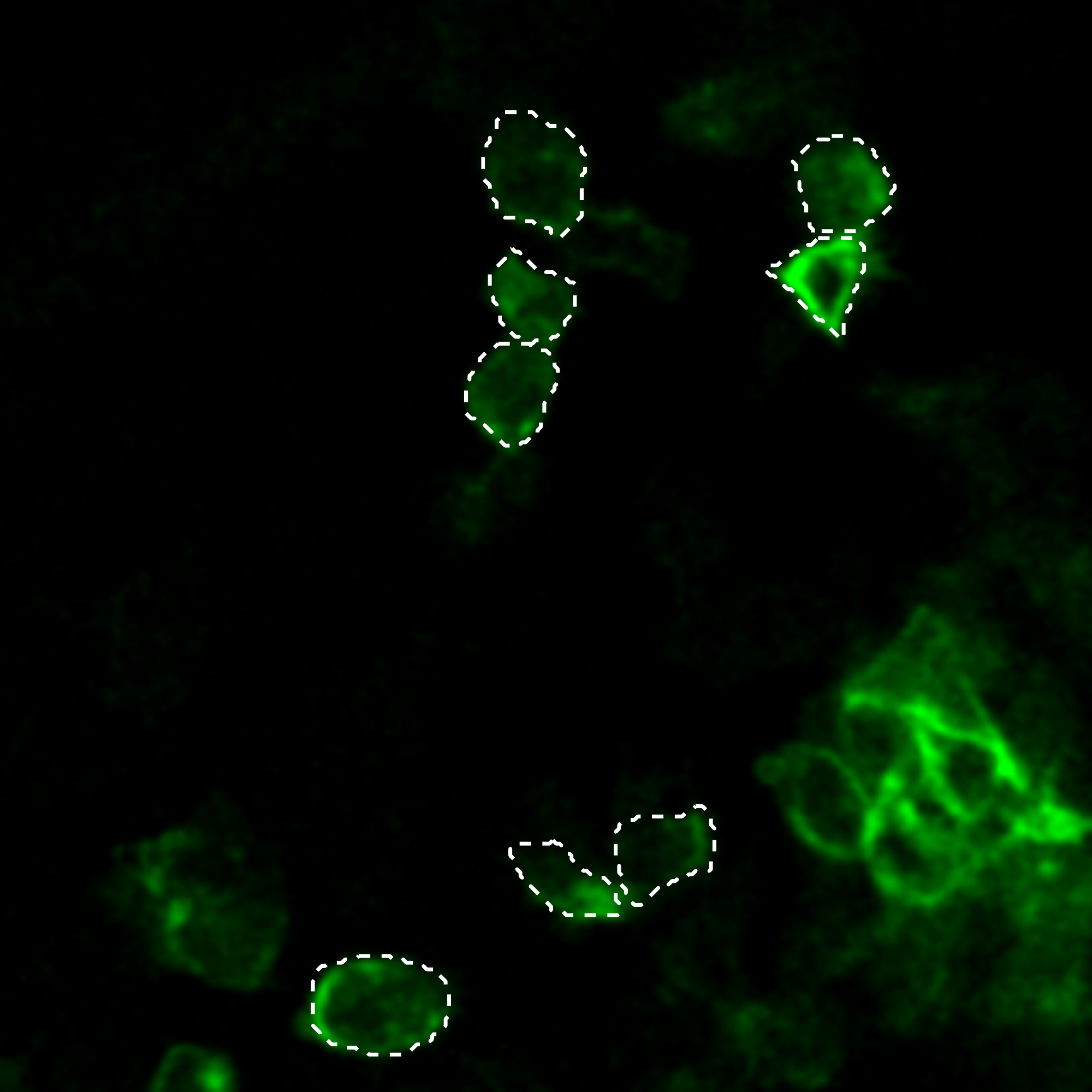}}
  \smallskip\centerline{\begin{footnotesize}
  a. Input and ground truth
  \end{footnotesize}}\medskip
	\centerline{\includegraphics[width=\imgwidth]{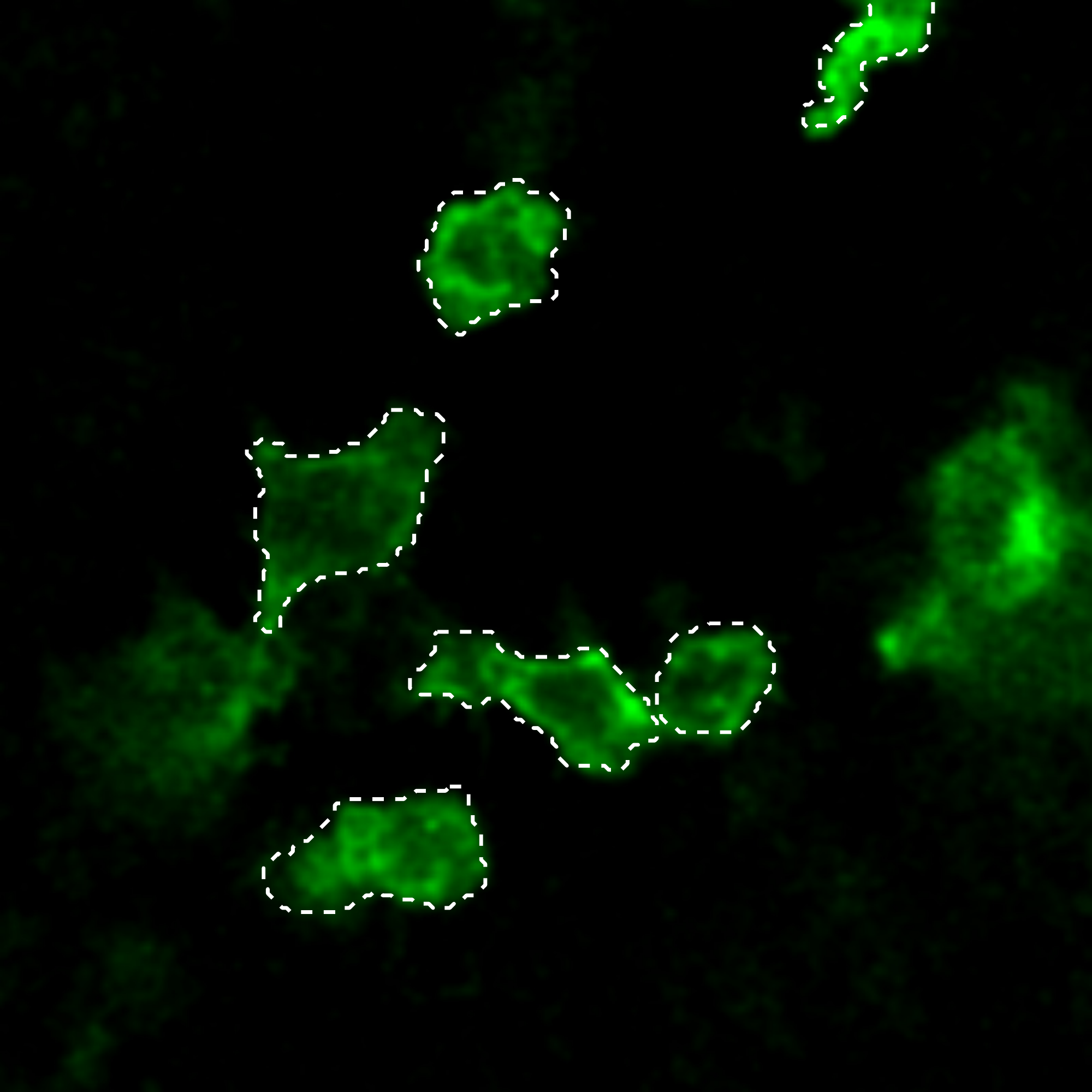}}
\end{minipage}
\hfill
\begin{minipage}[b]{\minipagewidth}
  \centering
  \centerline{\includegraphics[width=\imgwidth]{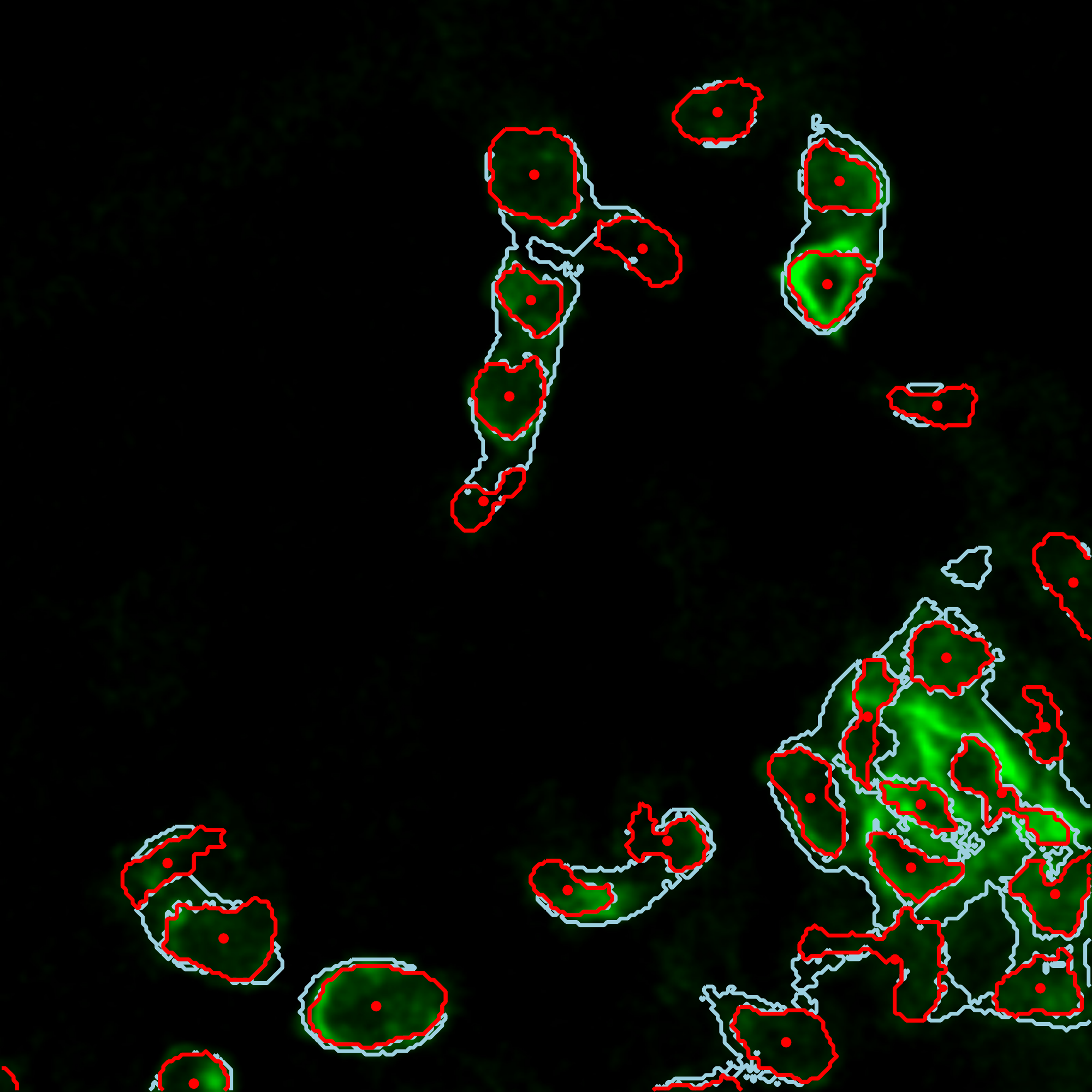}}
  \smallskip\centerline{\begin{footnotesize}
  b. ArgMax classification
  \end{footnotesize}}\medskip
    \centerline{\includegraphics[width=\imgwidth]{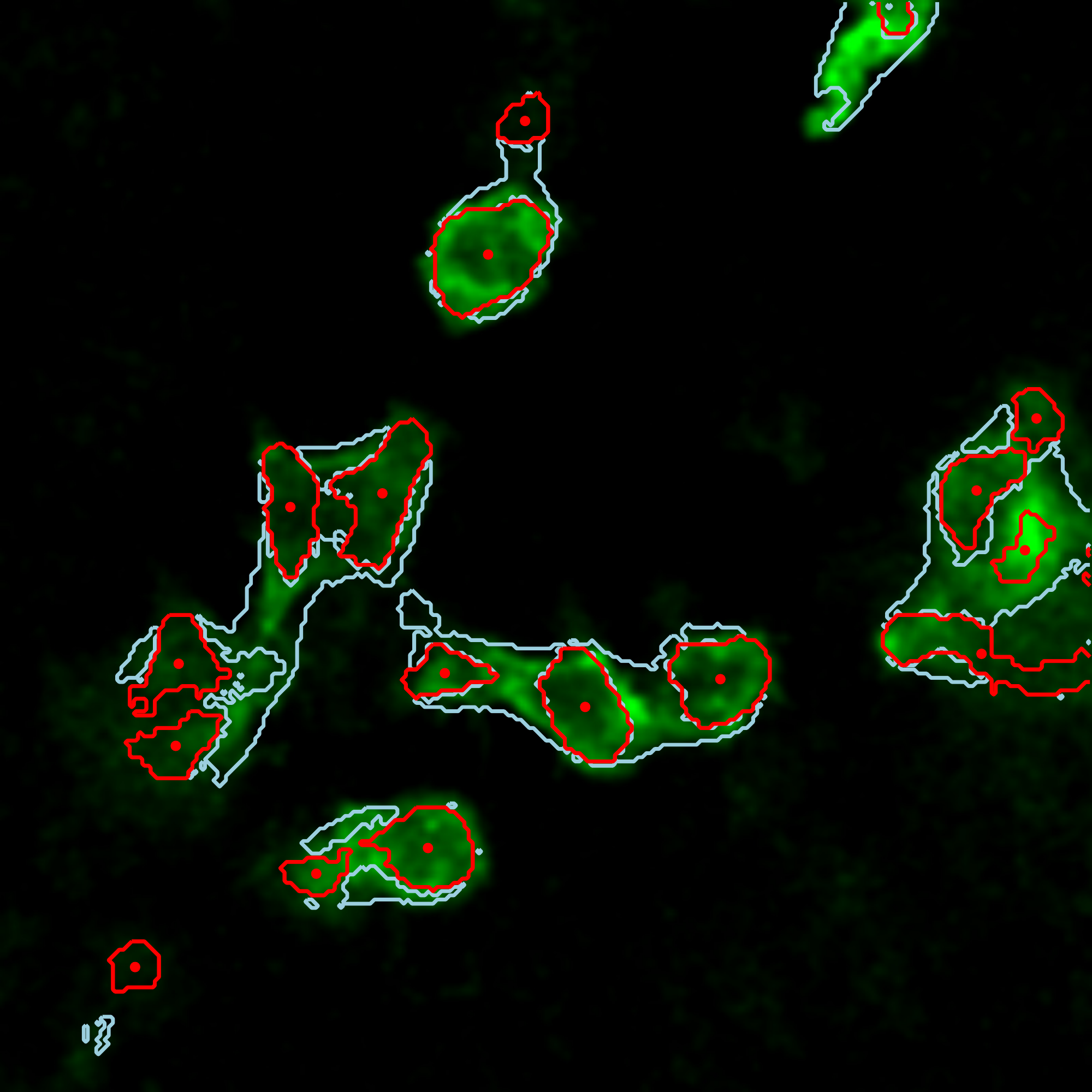}}
\end{minipage}
\hfill
\begin{minipage}[b]{\minipagewidth}
  \centering
  \centerline{\includegraphics[width=\imgwidth]{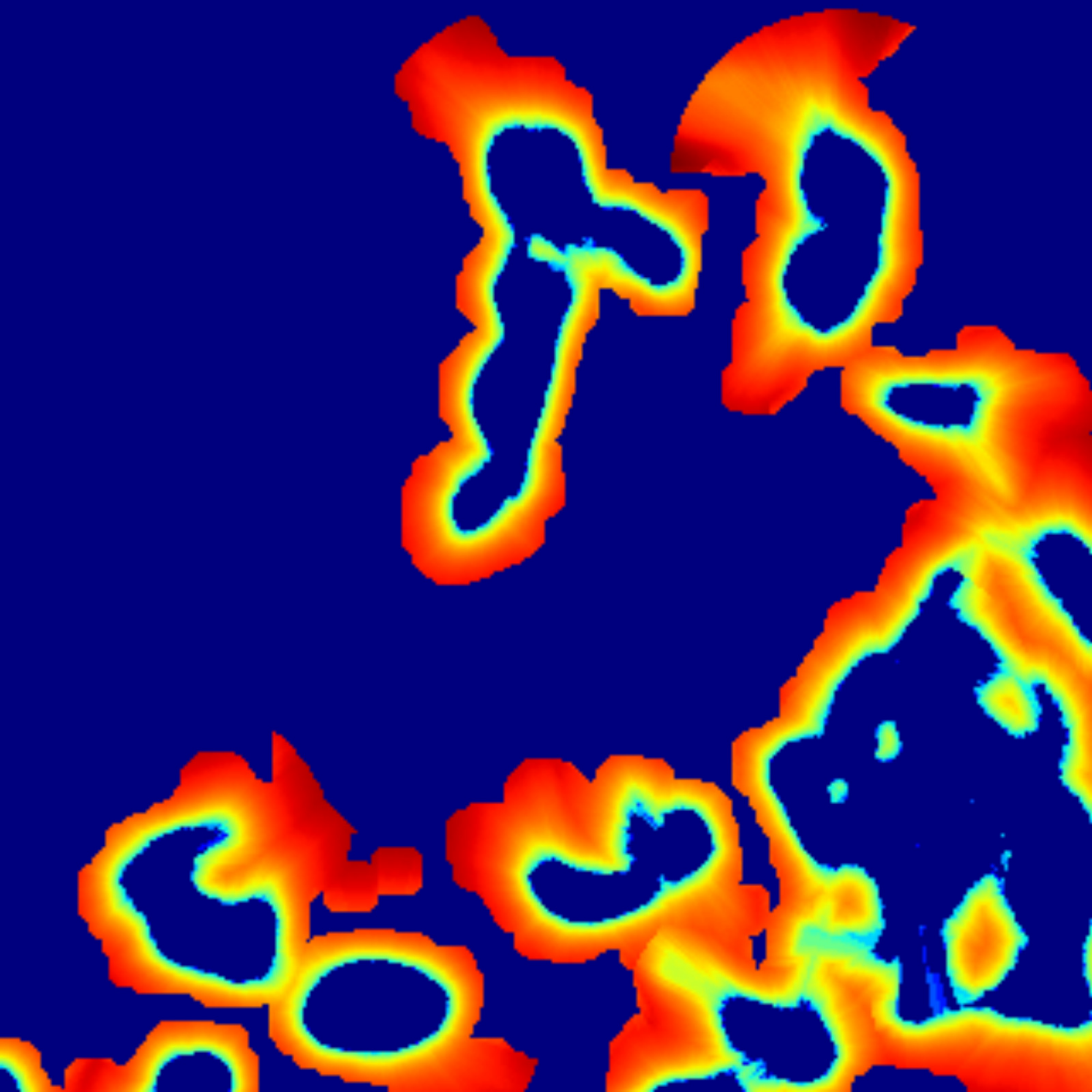}}
  \smallskip\centerline{\begin{footnotesize}c. $\min\int s_e-s_i$\end{footnotesize}}\medskip
    \centerline{\includegraphics[width=\imgwidth]{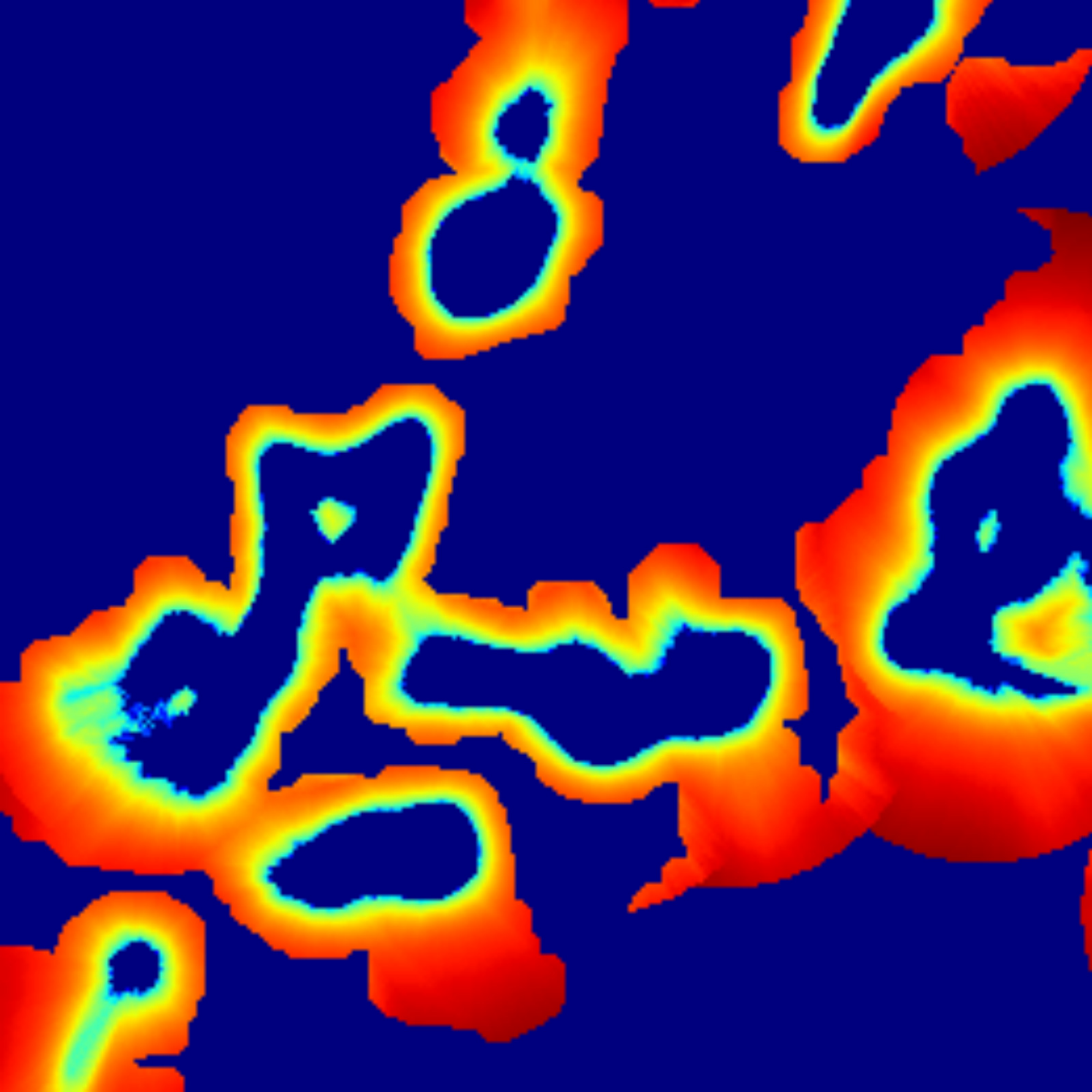}}
\end{minipage}
\hfill
\begin{minipage}[b]{\minipagewidth}
  \centering
  \centerline{\includegraphics[width=\imgwidth]{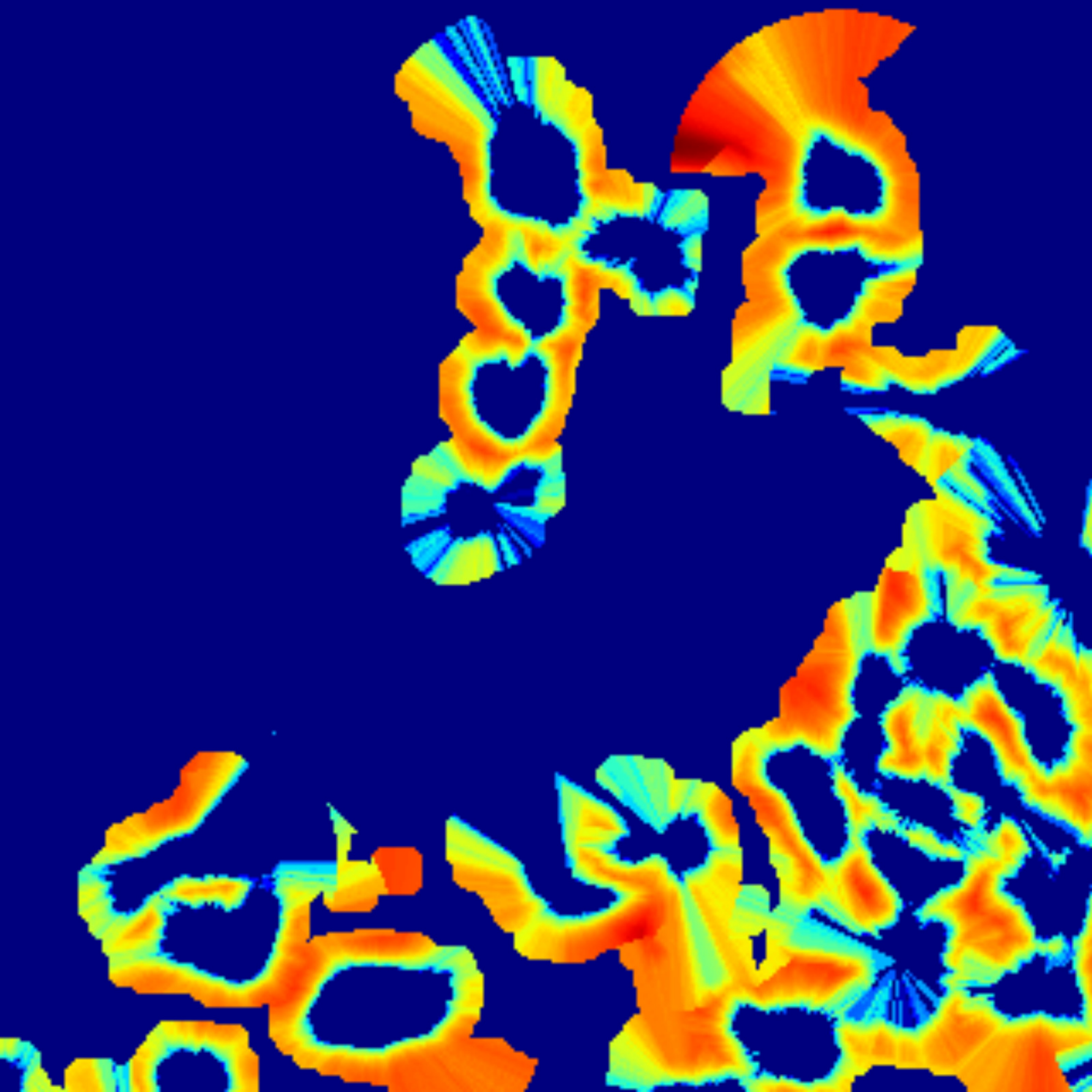}}
  \smallskip\centerline{\begin{footnotesize}d. $\min\int s_b-s_i$\end{footnotesize}}\medskip
    \centerline{\includegraphics[width=\imgwidth]{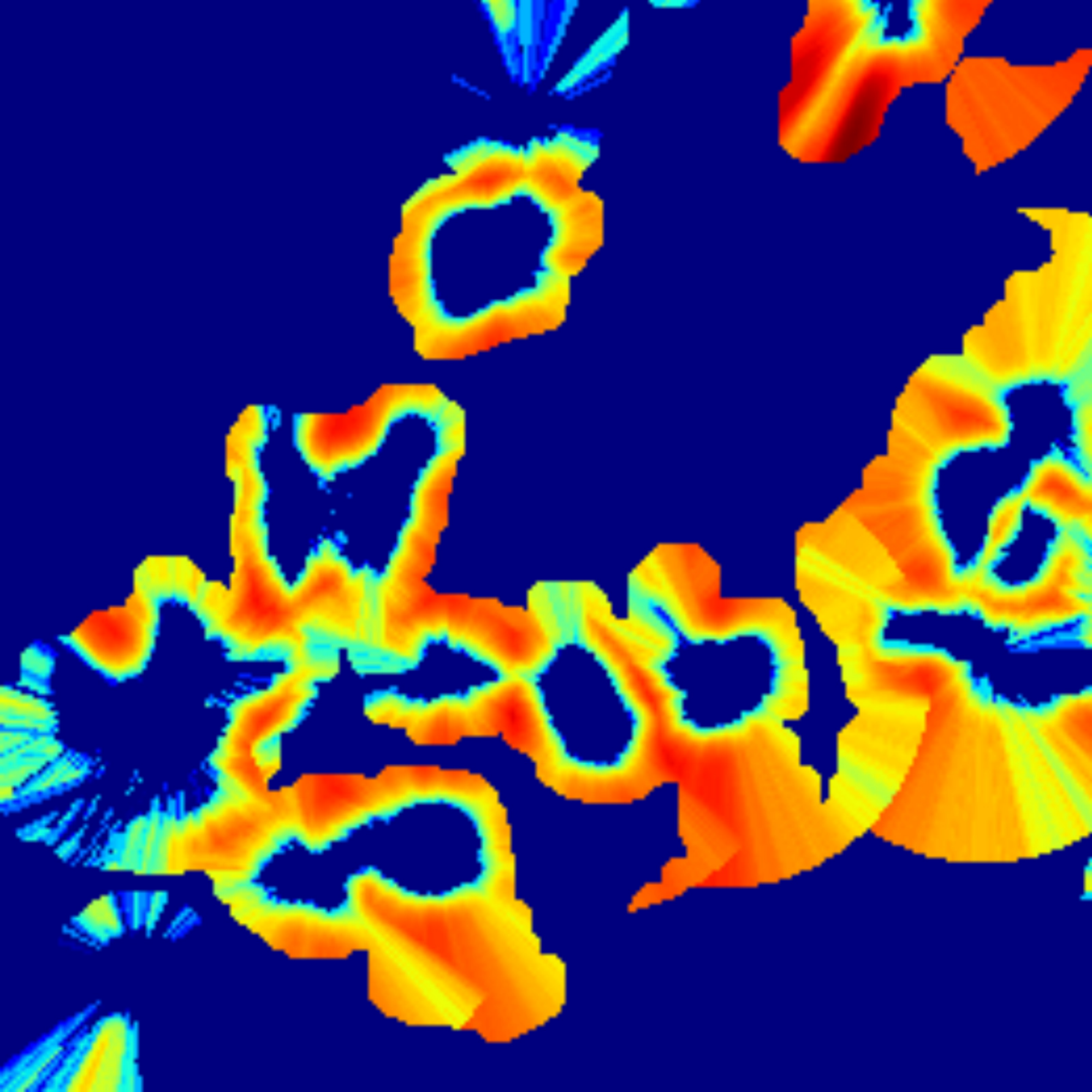}}
\end{minipage}
\hfill
\begin{minipage}[b]{\minipagewidth}
  \centering
  \centerline{\includegraphics[width=\imgwidth]{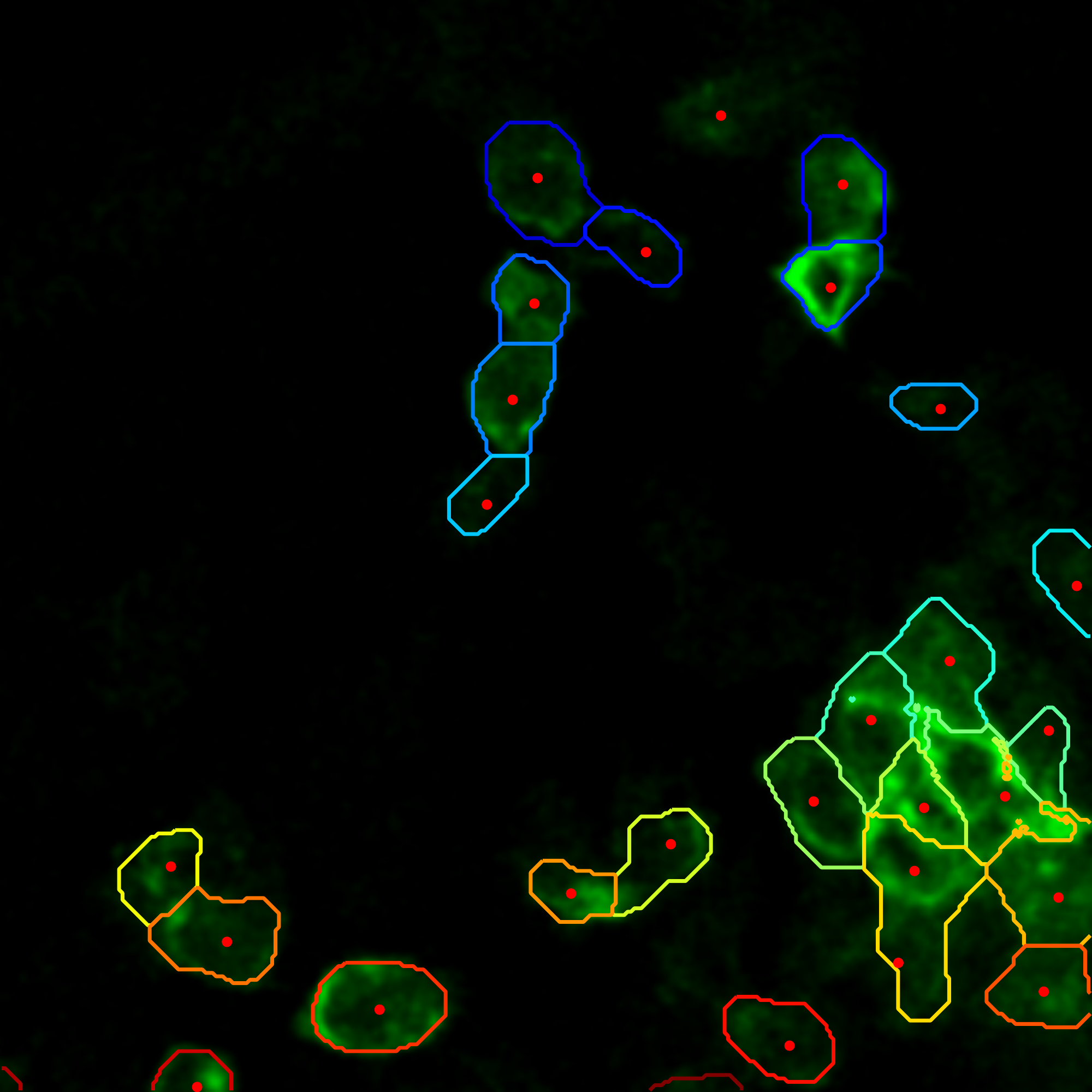}}
  \smallskip\centerline{\begin{footnotesize}e. Final result\end{footnotesize}}\medskip
    \centerline{\includegraphics[width=\imgwidth]{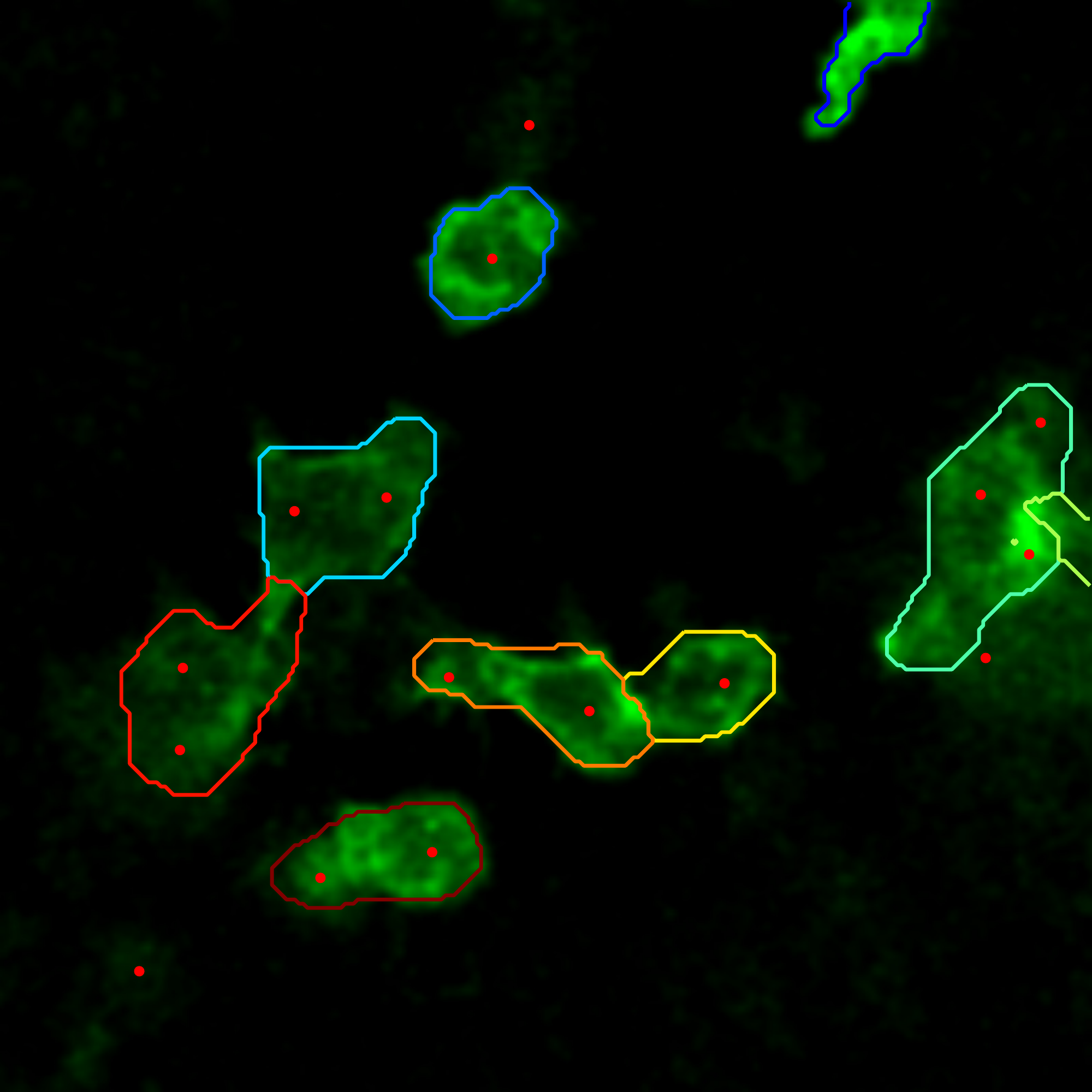}}
\end{minipage}
\caption{a) Ground truth, b) MAP estimator of the pixels class from the random ferns scores, red corresponds to the interior class and light blue to the boundary class (exterior class not displayed), c) heatmap of minimal line integral of $s_e-s_i$, d) heatmap of minimal line integral of $s_b-s_i$, e) final result with graph cut segmentation. The red dots in b and e correspond to the seeds extracted from the interior score distribution (best viewed in color and by zooming on screen).}
\label{fig:final}
\vspace{-.3cm}
\end{figure*}
}

\begin{document}
%\ninept
%
\maketitle
\begin{abstract}
The progress in imaging techniques have allowed the study of various aspect of cellular mechanisms. To isolate individual cells in live imaging data, we introduce an elegant image segmentation framework that effectively extracts cell boundaries, even in the presence of poor edge details. Our approach works in two stages. First, we estimate pixel interior/border/exterior class probabilities using random ferns. Then, we use an energy minimization framework to compute boundaries whose localization is compliant with the pixel class probabilities. We validate our approach on a manually annotated dataset.
%In particular, we are interested in cellular migrations during mouse embryo gastrulation.
% We show that those probabilities carry insightful information about the pixel class distribution in the images.
\end{abstract}

\begin{keywords}
cell segmentation, fluorescent microscopy, random ferns, graph-cuts
\end{keywords}

\rms
\section{Introduction / Overview}
\label{sec:intro}
Embryo morphogenesis relies on coordinated cell movements and tissue reorganization to allow correct shaping. Progress in embryo culture and live imaging techniques has allowed direct observation of cellular rearrangements in embryos from various species, including those with internal development~\cite{Nowotschin2014}. Important insight has been obtained through qualitative analysis of live imaging data, but quantitative automated analysis remains a bottleneck. The specific question addressed here is the cellular mechanisms of mesoderm migration during mouse embryo gastrulation~\cite{Arnold2009}. 
%Gastrulation occurs at embryonic day e6.25, when cells of the epiblast ingress the primitive streak, undergo an epithelial-mesenchymal transition to become mesoderm or endoderm, and migrate into the space between the epiblast and the visceral endoderm (Arnold 2009).
To look at cell shape changes of the nascent mesoderm after ingression, we examine Brachyury-Cre; mTomato/mGFP embryos between e6.75 and e7.5 by confocal microscopy. Cells expressing Brachyury that have gone through the streak and are populating the embryos as migrating mesoderm have green membranes, while the rest of the embryo has red membranes. To ensure optimal embryo survival, it is best to avoid multiple colors imaging~\cite{Lou2014}, and we have thus favored a membrane marker. The goal is to track cell movements to build a map of cell trajectories depending on time and place of ingression. 
\begin{figure}[t]
\begin{minipage}[b]{.48\linewidth}
  \centering
  \centerline{\includegraphics[width=4.0cm]{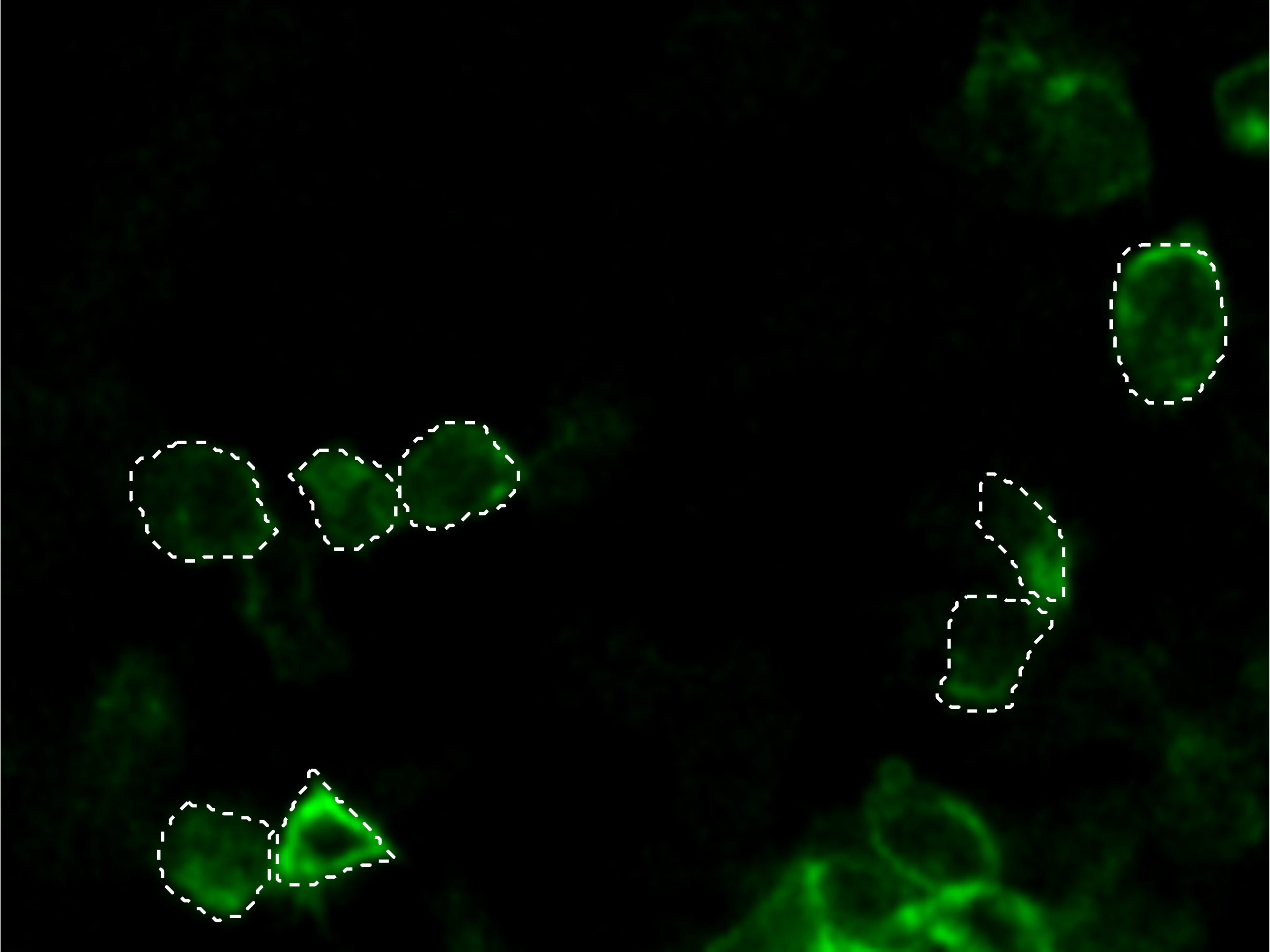}}
%  \vspace{1.5cm}
  \centerline{{\footnotesize (a) Input and ground truth}}\medskip
\end{minipage}
\hfill
\begin{minipage}[b]{0.48\linewidth}
  \centering
  \centerline{\includegraphics[width=4.0cm]{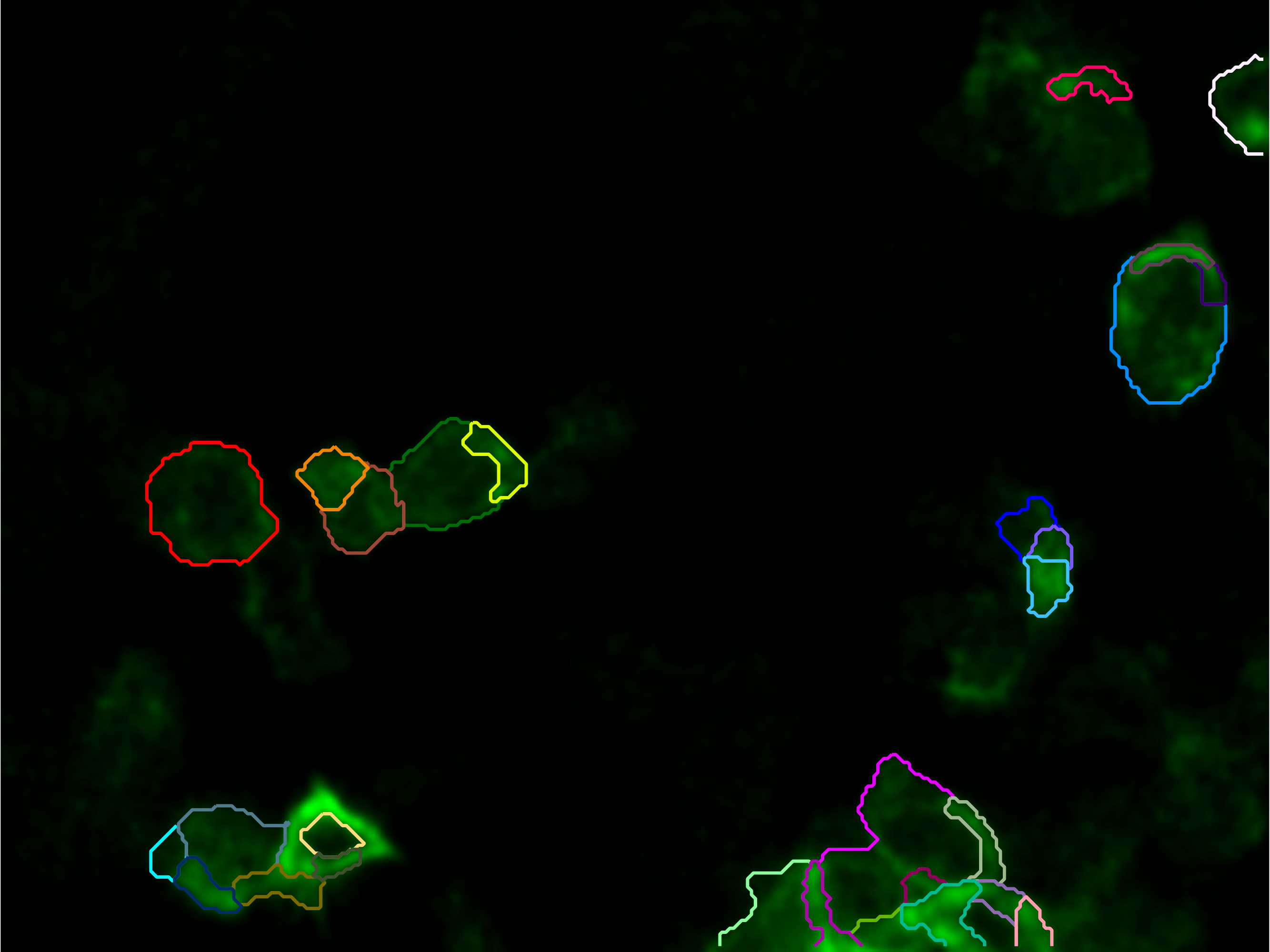}}
%  \vspace{1.5cm}
  \centerline{{\footnotesize (b) Mean-Shift~\cite{MeanShiftPAMI2002}}}\medskip
\end{minipage}

\begin{minipage}[b]{.48\linewidth}
  \centering
  \centerline{\includegraphics[width=4.0cm]{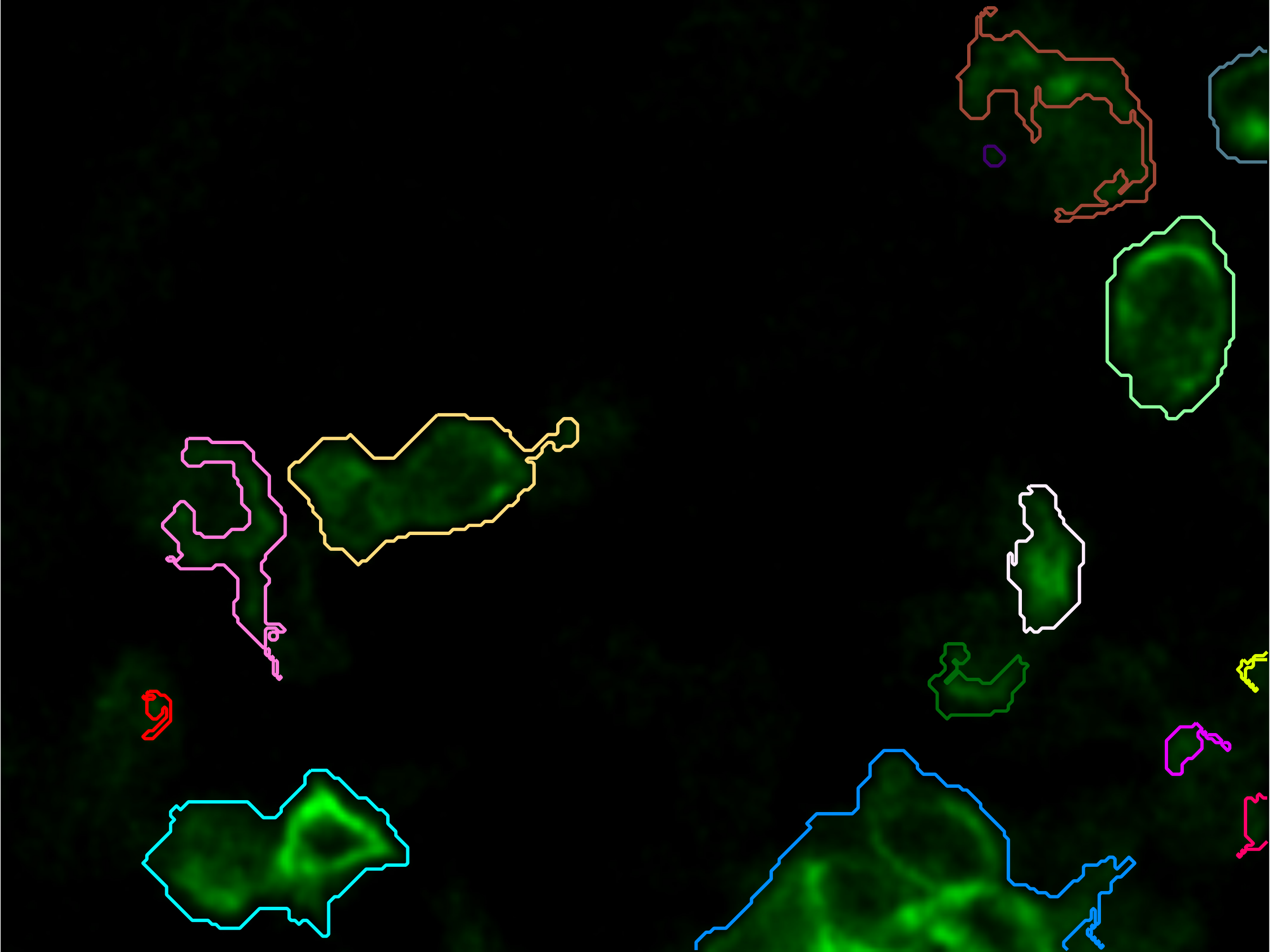}}
%  \vspace{1.5cm}
  \centerline{{\footnotesize (c) Felzenszwalb et al.~\cite{FelzenszwalbIJCV2004}}}\medskip
\end{minipage}
\hfill
\begin{minipage}[b]{0.48\linewidth}
  \centering
  \centerline{\includegraphics[width=4.0cm]{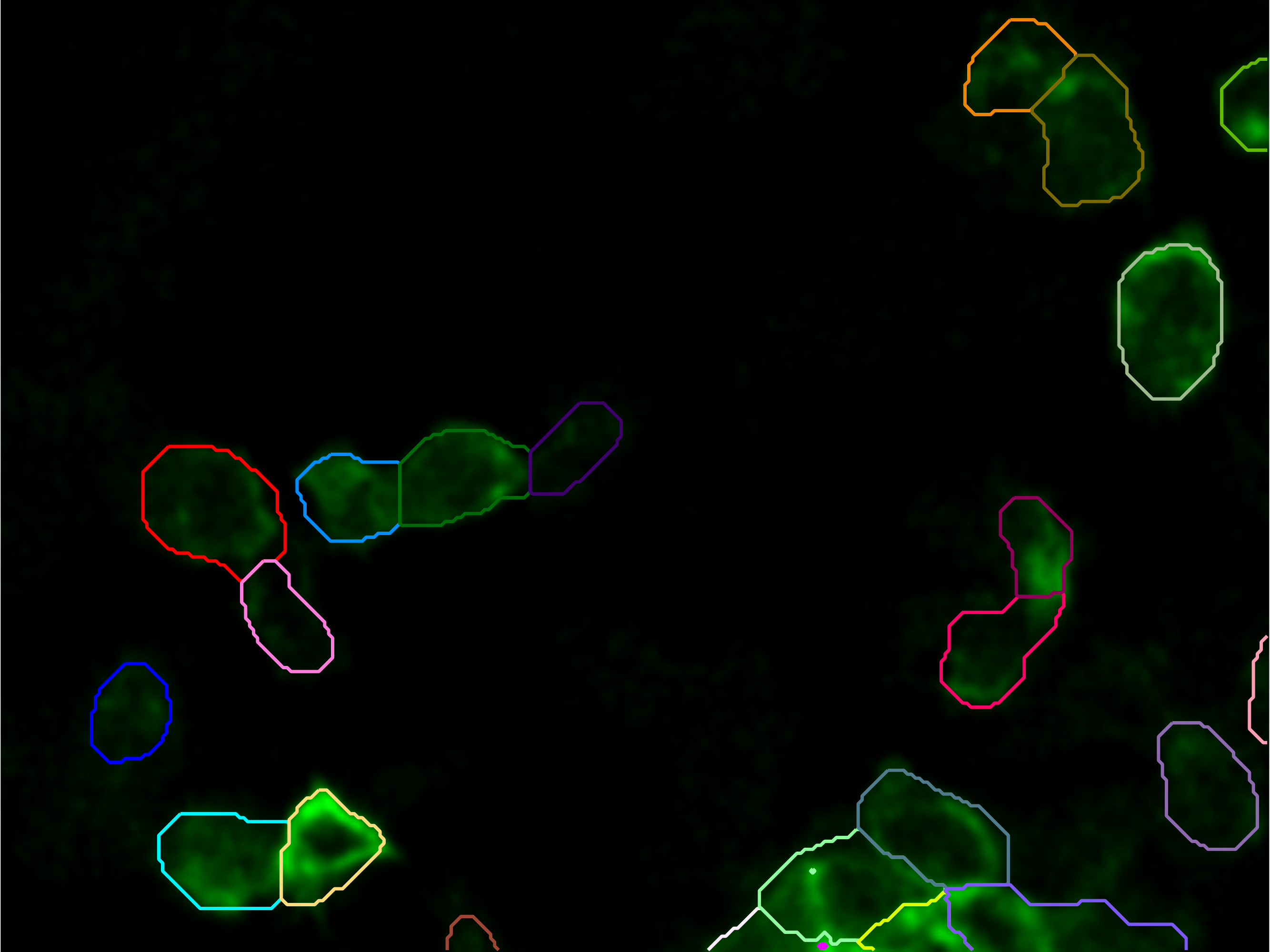}}
%  \vspace{1.5cm}
  \centerline{{\footnotesize (d) Our result}}\medskip
\end{minipage}
\vspace{-.2cm}
\caption{ {\footnotesize Comparison of segmentation results (best viewed in color on screen).}}
\label{fig:comp}
\vspace{-.45cm}
\end{figure}

As a preliminary step to cell movement analysis, our work focuses on cell detection and segmentation. The images collected using fluorescence microscopy exhibit many characteristics that make segmentation challenging. These include limited spatial resolution and contrast, resulting in poor membrane details. Specifically, as can be observed in Fig.~\ref{fig:comp}.a, the fluorophores do not strictly concentrate along the cell membranes in our dataset~\cite{dataset}. In contrast to images studied in~\cite{Fernandez2010, Khan2895, Mosaliganti2012}, this makes the border between two adjacent cells difficult to isolate, even visually. Moreover, the inner textures of distinct cells present quite similar statistics, making region merging strategies inappropriate as long as they do not use edge information.  This is in contrast with natural images, in which the objects to segment are characterized by distinct inner textures and color, and can therefore be effectively segmented using superpixel merging techniques~\cite{HierarchicalContourRegionPAMI2011}.

Fig.~\ref{fig:comp} illustrates two approaches that are widely employed for image segmentation. The graph-based method of Felzenszwalb and  Huttenlocher merges the regions in a greedy manner, using a minimum spanning-tree to measure the pixels uniformity in a region and compare it to border transitions~\cite{FelzenszwalbIJCV2004,CausalICIP2013}. The Mean Shift (MS) algorithm offers an alternative popular clustering framework. MS represents pixels in the joint spatial-range domain by concatenating their spatial coordinates and intensity values into a single vector. This method then assigns each pixel to a local maxima of the statistical distribution of the pixels in this domain, using a gradient-descent process~\cite{MeanShiftPAMI2002}. We observe in Fig.~\ref{fig:comp} that none of these approaches succeeds in segmenting adjacent cells. Hence, they are not able to capture the semantic knowledge required to distinguish individual cells within cells aggregate.
% It attempts to partition image pixels into components such that the resulting segmentation is neither too coarse (non-uniform regions) nor too fine (uniform regions oversegmentation).

Among the approaches proposed in the literature to address semantic segmentation problems,~\cite{BoykovBasedInteractiveICIP2010} and~\cite{PointsPriorICIP2006} have respectively considered an interactive framework or a prior discriminative description of the object to segment. In the context of microscopy,~\cite{RohdeCMU2013} have defined such prior models based on templates, learned in a supervised manner. In super-resolution localization microscopy and in MRI,~\cite{SLRM-ICIP2014} and~\cite{SVMCassifierICIP2013} respectively rely on density estimation or SVM texture features classification to differentiate structures of interest. Those approaches are however only relevant when strong appearance priors exist about how the object to segment differs from its environment. This is not the case in our dataset, where the shape of the cells is subject to significant variability, and where the environment of each cell is composed of quite similar other cell patterns.

In cases where the object appearance is not discriminant, training appropriate edge detectors appears to be a natural approach~\cite{MairalECCV2008, CassifierWatershedICIP2013}. The work in~\cite{CassifierWatershedICIP2013} is of particular interest. It has been proposed in the context of neurons reconstruction, using electron microscopy. It combines a pixel-level membrane probability estimator with a conventional watershed algorithm to segment regions that are likely to be closed by a membrane. In practice however, the membrane probability map presents too many local minima, which leads to an oversegmented partition. To address this problem, a so-called boundary classifier is trained to control the merging of adjacent regions, based on the statistics of boundary and region pixels. The main drawback of this approach is that the boundary classifier is trained directly on the output of the watershed stage, thereby requiring training adjustment when the watershed thresholds are tuned. Moreover, the contours defined in the first step, strictly based on the membrane detector, can only be removed in the second step, without being corrected based on the observed region pixel statistics.

To circumvent those limitations, we propose to adopt an approach that does not consider edge- and inside- pixels sequentially, but instead considers them jointly. In an initial stage, our approach learns how interior pixels differ from background or border pixels. It then adopts a global energy minimization framework to assign cell-representative labels to pixels, based on their posterior interior/border/exterior class probabilities. Considering explicitly a class of pixels lying on borders between adjacent cells is critical since the main problem encountered by previous works on our dataset consists in splitting cellular aggregates into individual cells (see Fig.~\ref{fig:comp}). Formally, we use a semi-Naive Bayesian approach to estimate, in each pixel, the probabilities that this pixel lies inside a cell, on a boundary between adjacent cells, and in the background. We have chosen semi-Naive Bayesian estimation because it has been shown to be accurate and offer good robustness and generalization properties in many vision classification tasks~\cite{Ozuysal2010,parisotACIVS2013}. This last point is important since the manual definition of cell contour ground-truth is generally considered as a tedious task, which practically limits the number of available training samples. Regarding the subsequent energy-minimization framework, we rely on the fast approximate minimization with label costs introduced by Delong et al.~\cite{Boykov2}, based on the seminal work of Boykov et al.~\cite{Boykov1}. In final, our work appears to be an elegant and effective solution to exploit posterior interior/border/exterior probability maps in a segmentation context.

The rest of the paper is organized as follows. Section~\ref{sec:rf} introduces our semi-Naive Bayesian probability vector estimator. Section~\ref{sec:gcut} describes the energy minimization labelling framework. Section~\ref{sec:results} validates our approach, and Section~\ref{sec:ccl} provides some concluding comments.

%ground truth has been manually annotated, tedious task. Induce a very small amount of annotation for training.  this should be considered when selecting a classification tools

\rms
\section{Pixel class probability estimation}
\label{sec:rf}
This section explains how to assign interior/border/exterior class probabilities to a pixel, based on the observation of its neighborhood. Following many successful recent works~\cite{Ozuysal2010,ClassificationZisserman,XtraTrees}, we use randomized sets of binary tests to characterize the different classes of point neighborhoods. 

In practice, the point neighborhood is defined by a small square window of radius $l$ and of size $(2l+1)^2$ centered around the pixel of interest. Each binary test compares the intensity of two pixels, and is set to 1 when the first is larger than the second, and to 0 otherwise. The pixel positions of each test are drawn uniformly at random within the square window. 
The approach considers $N \in \mathbb{N}$ sets of $S \in \mathbb{N}$ binary tests that are randomly selected, to define $N$ flat structures, named \emph{ferns}. 

As in~\cite{Ozuysal2010}, let $C \in \mathcal{C}$ denote the random variable that represents the class of an image sample, and $\mathcal{C} = \{c_i: 0 < i \leq H\}$ be the set of $H =3$ interior/border/exterior classes. Given the ensemble of $N$ ferns $\mathcal{F} = \{F_k \in \{0,1\}^S: 1\leq k \leq N\}$, where $F_k$ denotes the $k^{\textrm {th}}$ fern,  we are interested in estimating the posterior probabilities $P(C=c_i | F_1,\,\cdots,F_N)$. 
%:
%\begin{equation}
%\hat{c} = \arg\max_{c_i} P(C=c_i | F_1,\,\cdots,F_N).
%\label{eq:map-estimate}
%\end{equation}
If we admit a uniform prior with $P(C=c_i)=1/H$ for $1\leq i\leq H$, Bayes' formula yields:

\begin{footnotesize}
\vspace{-.3cm}
\begin{equation}
P(C=c_i | F_1,\,\cdots,F_N) \propto P(F_1,\,\cdots,F_N | C = c_i).\phantom{\prod}
\label{eq:joint}
\end{equation}
\vspace{-.3cm}
\end{footnotesize}

\indent Learning and handling the class conditional joint probability in (\ref{eq:joint}) is not feasible for large $N \times S$ products since it would require to compute and store $2^{NS}$ entries for each class. To keep the conditional probabilities tractable while accounting for some binary tests dependencies, the semi-naive Bayesian approach proposed in~\cite{Ozuysal2010} assumes independence between the ferns, but accounts for dependencies between the binary tests belonging to the same fern. The joint conditional probability is approximated by:
%\begin{footnotesize}
%\vspace{.1cm}
\begin{equation}
\label{eq:classcond}
 P(F_1,\,\cdots,F_N | C = c_i) \simeq \prod_{k=1}^N P(F_k| C = c_i),
\end{equation}
%\vspace{-.4cm}
%\end{footnotesize}
where the class conditional distribution of each fern is simply learned based on the accumulation of the training samples observations, as detailed in~\cite{Ozuysal2010}.

When the number of ferns is large, the product in (\ref{eq:classcond}) may cause computational underflow. Hence, in general, one defines the score 
%\begin{footnotesize}
%\vspace{.1cm}
\begin{equation}
s_c = \log\left(\sigma_c\right) = \sum\limits_{k=1}^{N}\log\left(P\left( F_k \left| C = c \right.\right)\right),
\label{eq:score}
\end{equation}
%\vspace{-.4cm}
%\end{footnotesize}
The scores extracted with the random ferns provide an interesting insight about the class distribution of pixels within the image. 
In a conventional classification framework, a pixel class MAP estimate $\hat{c}$ is defined by:

\begin{footnotesize}
\begin{equation}
\hat{c} = \arg\max_{c\in \mathcal{C}} P(C=c | F_1,\,\cdots,F_N) \simeq \argmax\limits_{c\in \mathcal{C}} s_c.\phantom{\prod}
\label{eq:map-estimate}
\end{equation} 
\end{footnotesize}
\indent In our segmentation problem, however, the MAP does not define accurately the cell boundaries, see Fig.~\ref{fig:final}.b. Therefore, we turn to a global energy-minimization, build upon the ferns scores, to derive an appropriate segmentation. In what follows, when we refer to the ferns scores, we consider them normalized, \ie $\widetilde{s_c} = s_c/\left(\sum_cs_c\right)$, although we will abuse the notation $s_c$ for clarity.

\rms
\section{Class compliant energy minimization}
\label{sec:gcut}

The global energy minimization framework introduced in~\cite{Boykov1, Boykov2} is used to assign cell-representative labels to pixels, based on their posterior interior/border/exterior class probabilities. Given a set of $n$ labels $\mathcal{L} = \left\{ 1,{}\cdots{},n\right\}$, 
we are looking for a pixel-to-label assignment $f$ that minimizes the energy

\begin{footnotesize}
\begin{equation}
E\left(f\right) = \sum\limits_{p \in \mathcal{P}} D_p\left(f_p\right)
 + \sum\limits_{(p,q)\in\mathcal{N}}W(p,q)\left(1-\delta(f_p,f_q)\right) + \sum\limits_{l\in\mathcal{L}}h_l\left(f\right),
\label{eq:boykov}
\end{equation}
\vspace{-.3cm}
\end{footnotesize}

\noindent where $\delta$ is the Kronecker delta, $\mathcal{P}$ stands for the set of pixels and $\mathcal{N}$ for a set of pairs of interacting pixels. As detailed below, the first term, $D_p\left(f_p\right)$, is called data fidelity and measures the cost to associated each pixel $p$ to its label $f_p$. The second term, $W(p,q)$, regularizes the label assignment by penalizing the assignment of distinct labels to interacting pixels $p$ and $q$ in a graph structure $\left(\mathcal{P}, W\right)$. Finally, as detailed in~\cite{Boykov2}, the last term, $h_l\left(f\right)$, introduces a cost when $f$ assigns the label $l$ to at least one pixel. The graph structure penalizes local inconsistencies of the labels while the label cost penalizes having to many different labels globally.
%to insert each label within our label set

We initialize the label set so that each cell is represented by at least one label. To do so, we extract a number of cell-representative {\em{seeds}}. In practice, each seed corresponds to the center of a connected set of pixels whose interior score lies above a threshold. To circumvent the threshold selection issue, and to adapt the seed definition to the local image contrast, we consider a decreasing sequence of thresholds. Large thresholds result in small segments, that progressively grow and merge as the threshold decreases. Among those segments, we only keep the largest ones whose size remains (significantly) smaller than the expected cell size. This might result in multiple seeds per cell, as depicted by red dots in Fig.~\ref{fig:final}.b and e. A unique label is then attached to each seed, adding one virtual label for the background. The fact that a single cell induces multiple seeds, and thus multiple labels, is not dramatic since the subsequent energy-minimization tends to filter redundant labels. 

To obtain a label assignment that is compliant with the class probabilities obtained in Section~\ref{sec:rf}, we define the cost functions in~(\ref{eq:boykov}) upon the ferns scores:
%We now define each term in~(\ref{eq:boykov}):

%The main advantage of this energy minimization framework is that it can merge seeds to a common labels when this proves to be useful.
\includelargeimage
%With the definition of such seeds, the watershed algorithm is a natural choice to compute a segmentation~\cite{CassifierWatershedICIP2013}. However, this algorithm is not able to correctly handle the pictures under study because 1) nor the intensity, nor the gradient landscape are suited to extract accurate boundaries 2) some cells are spanned by more than one seed which constitute a severe drawback. 
%\begin{itemize}
\noindent $\bullet$~\emph{\textbf{the data fidelity}} of assigning a pixel $p$ to a seed label $f_p$ builds on two complementary signals because we want the cost to increase largely when the path from a pixel to a seed crosses a cell border, whether this border is between the cell and the background or between two cells. Hence,

\begin{footnotesize}
\vspace{-.3cm}
\begin{equation}
D_p(f_p) = \int_\rho \max\left(0,\, s_e(\rho)-s_i(\rho), \,s_b(\rho)-s_i(\rho)\right),
\label{eq:DF}
\end{equation}
\vspace{-.3cm}
\end{footnotesize}

\noindent where  $s_i$, $s_b$ and $s_e$ correspond to the fern scores for the interior, the boundary and the exterior classes respectively, and $\rho$ is the set of pixels along the line connecting pixel $p$ to the seed associated to $f_p$~\cite{bressenham}. The $\max$ operator is used to penalize the allocation of $p$ to $f_p$ only when $\rho$ crosses a border, \ie $s_e> s_i$ or $s_b>s_i$. As depicted in Fig.~\ref{fig:final}.c and d, the signal $s_e-s_i$ indeed peaks for borders between the cells and the background while the signal $s_b-s_i$ peaks for borders between two cells.

\noindent $\bullet$~\emph{\textbf{the data fidelity}} of assigning a pixel $p$ to the background is the minimal exterior score integral computed over the set of lines $\Gamma_p^d$, each line originating in $p$, and having a length $d$. Hence, 
%\begin{footnotesize}
\begin{equation}
D_p(f_p=e) = \min\limits_{\rho\in\Gamma_p^d} \int_\rho s_e(\rho).
\end{equation}
%\end{footnotesize}
 %In practice, we sample a set of $8$ basic directions and select the minimal value.
\noindent $\bullet$~\emph{\textbf{the graph edge weight}} $W(p,q)$ between interacting pixels, that are adjacent pixels on an 8-neighborhood connectivity, is computed using a sigmoid function as  $$W(p,q) = 1-\frac{1}{1+\alpha_we^{-\beta_w M(p,q) } },$$ where $M(p,q)$ is defined by
$$M(p,q) = \max_{k\in\left\{p,q\right\}} \left(min\biggl(s_b(k)-s_i(k), s_b(k)-s_e(k)\biggr)\right).$$
Doing so, the edge weight is low (high), allowing (discouraging) neighboring pixels to have different labels, when the probability of having a boundary at pixels $p$ or $q$ is high (low). Values for $\alpha_w$ and $\beta_w$ are not critical and chosen empirically.
%
%there is a high (low) probability of a boundary around the pixels $p$ and $q$. 
%\end{itemize}

Minimizing (\ref{eq:boykov}) is NP-hard. We compute an approximate solution efficiently using graph-cuts, with $\alpha$-expansions, as described in~\cite{Boykov2}. This energy minimization framework is particularly well suited to our problem because it may account for multiple seeds spanning the same cell, as opposed to classical watershed approaches~\cite{CassifierWatershedICIP2013}.% Moreover, as demonstrated in Section~\ref{sec:results}, the regions extracted are in good agreement with the ground truth.
% computes a near-optimal label distribution across the set of pixels $\mathcal{P}$ according to the ferns scores. The method

\rms
\section{Experimental results}
\label{sec:results}
We validate our segmentation framework on a sequence of images with manually annotated ground truth, publicly released~\cite{dataset}\footnote{To favor reproducible research, our code will also be made publicly available at camera ready submission}.% composed of $25$ time points with $20$ slices per time point. 

%While different pre-processing of the input pictures were considered, we observe that the best results are obtained by applying only a small Gaussian filter with $\sigma=1$ to reduce the noise of the acquisition process.

%The $3$ classes of pixels are defined by applying morphological operation on the annotated ground truth. The interior class is obtained with binary erosion with a circle of radius $5$ as the structuring element. The exterior class is defined by the set of points in between the regions obtained by applying morphological dilation with circle of radius $3$ and $5$ as the structuring elements. Finally, the boundary class, which contains points lying on the border between cells, is defined by considering the points obtained as exterior points for at least $2$ different annotated cells. We observe that adding a class for pixels lying on the border between the cells and the background did not improve the quality of the segmentation.

To define our training set based on the manually annotated cell contours (white dashed contours in Fig.~\ref{fig:final}.a), we rely on morphological operations. Specifically, the interior class is set with binary erosion while the exterior class is set with binary dilation. The boundary class is composed of pixels lying on the exterior region of at least $2$ different cells. %We observe that adding additional classes did not improve the quality of the segmentation. 

Since the number of annotated cells is limited and because we enforce balanced classes for training, our training set is restricted to $1500$ pixels for each class.
%uniformly across the $24$ images manually annotated. This corresponds to the number of points available in the boundary class. Because of this strong constrain on the number of points available for training, we used $T=10$ tests per fern. 
%We consider each training frame $10$ times with different orientations to reduce the risk of overfitting on the training data. 
To increase the training set diversity and become invariant to rotation, we train the ferns on square windows that sample the image according to $10$ different orientations. Each fern involves 10 tests, and we use 200 ferns.
To measure the overall performance, we have run a $10$-fold cross-validation, and have measured a classification accuracy of $94\%$, with $1\%$ standard deviation.
%split the training set in $10$ non-overlapping subsets, then we recursively trained the ferns over $9$ of the subsets and measure the accuracy over the last one. 
%We observe that the classification rate saturates when using around $N=200$ ferns, with a classification accuracy around $94\%$ with a standard deviation of $1\%$.
% ($3$ different time points an $8$ stacks per time points)

%For the testing phase, we first filter only the relevant pixels by applying a classical Otsu thresholding {\color{red}(ADD REF)} and a morphological closing to reduce the time complexity. Overall, this selection is very loose and largely reduce the computational time without altering the classification since the rejected pixels are in the background. We compute the ferns scores for a frame size $l=15$ and then compute the data fidelity measures as well as the graph edge weight to be used in our segmentation phase. 

We have then tested our energy minimization framework on all the images available in the dataset\footnote{see \emph{http://perso.uclouvain.be/arnaud.browet/bioseg/results.html} for additional results.}$^,$\footnote{To avoid overfitting, each image has been segmented based on ferns trained exclusively from other images annotations.}.
%Again to reduce the time complexity, we only compute the data fidelity for points in a radius of $50$ pixels around each seeds. This value has been chosen according to the observed size of the annotated cells. Otherwise, the data fidelity is set to $\infty$. 
The parameters have been empirically selected as follows, $d=5$, $\alpha_w=40$ and $\beta_w=15$. Fig.~\ref{fig:final} presents some representative examples of segmentation, together with some insightful intermediate metrics.

Fig.~\ref{fig:final}.b depicts the segmentation resulting from the ferns only, using an argmax decision defined in (\ref{eq:map-estimate}). 

Fig.~\ref{fig:final}.c and d present the line integrals considered in equation (\ref{eq:DF}). Note that the integral values are only provided in pixels that lies within a 50 pixels distance from a seed.
%Let us note that we build the data fidelity terms only for pixels that are less than $50$ pixels away from a seed. 
This explains the particular landscape of Fig.~\ref{fig:final}.c and d. We observe that both metrics provide complementary information, delineating the cells either from the background or from an adjacent cells.

The last column in Fig.~\ref{fig:final} presents the segmentation resulting from our proposed ferns-based energy minimization. We observe that the regions extracted are in very good agreement with the ground truth. As depicted in the $1^{\textrm{st}}$ row of Fig.~\ref{fig:final}, our segmentation is able to accurately localize boundaries between touching cells. Moreover, our method is also able to merge multiple seeds within a unique region or to reject seeds situated in the background, as displayed in the $2^{\textrm{nd}}$ row of Fig.~\ref{fig:final}.
\rms
\section{conclusion}
\label{sec:ccl}
Our work has adopted an energy-minimization framework to segment cell images according to the cues provided by random ferns about the probability that each pixel is located within a cell or not.

Our framework is highly versatile, since the classes definition and the energy terms can account for any prior knowledge related to the problem at hand. Additionally, it is also interactive-friendly, in the sense that the seeds definition can easily be manually adjusted, if needed.
%for example through manual adaptation of seeds, in cases where those would be particularly badly identified. In future development of this work, we also aim to 3D reconstruct individual cells, based on those 2D segmentation, to follow cell shape changes, such as dynamics and polarization of projections, as well as to describe and quantify the extent of cell-cell contacts.

% Below is an example of how to insert images. Delete the ``\vspace'' line,
% uncomment the preceding line ``\centerline...'' and replace ``imageX.ps''
% with a suitable PostScript file name.
% -------------------------------------------------------------------------
%\begin{figure}[htb]
%
%\begin{minipage}[b]{1.0\linewidth}
%  \centering
%  \centerline{\includegraphics[width=8.5cm]{image1.eps}}
%%  \vspace{2.0cm}
%  \centerline{(a) Result 1}\medskip
%\end{minipage}
%%
%\begin{minipage}[b]{.48\linewidth}
%  \centering
%  \centerline{\includegraphics[width=4.0cm]{image3}}
%%  \vspace{1.5cm}
%  \centerline{(b) Results 3}\medskip
%\end{minipage}
%\hfill
%\begin{minipage}[b]{0.48\linewidth}
%  \centering
%  \centerline{\includegraphics[width=4.0cm]{image4}}
%%  \vspace{1.5cm}
%  \centerline{(c) Result 4}\medskip
%\end{minipage}
%%
%\caption{Example of placing a figure with experimental results.}
%\label{fig:res}
%%
%\end{figure}

%\vfill
%\pagebreak

\vfill
\pagebreak

%\newpage

\bibliographystyle{IEEEbib}
\bibliography{strings,refs}

\end{document}